\documentclass[11pt]{article}

\usepackage[a4paper,margin=1in]{geometry}
\usepackage{graphicx}
\usepackage{booktabs}
\usepackage{amsmath,amssymb}
\usepackage{hyperref}
\usepackage{cite}
\usepackage{caption}
\usepackage{subcaption}
\usepackage{float}
\usepackage{xcolor}
\usepackage{authblk}
\usepackage{enumitem}
\usepackage{algorithm}
\usepackage{algpseudocode}

\hypersetup{colorlinks=true, linkcolor=blue, citecolor=blue, urlcolor=blue}

\title{\textbf{HCIG: A Hierarchical Cross-Modal Incongruity Graph Network for Multimodal Sarcasm and Cyberbullying Detection}\\}

\author{
Bhavana Verma, Priyanka Meel, Dinesh Kumar Vishwakarma\\
\textit{Multimodal Data Analytics Research Laboratory}\\
\textit{Department of Information Technology}\\
\textit{Delhi Technological University}\\
\{bhavana.verma2905, priyankameel86, dvishwakarma\}@gmail.com
}

\date{}
\begin{document}
\maketitle

\begin{abstract}
\noindent
Multimodal sarcasm and cyberbullying detection remain challenging because the intended meaning often emerges from incongruity between textual and visual information rather than from either modality alone. Existing multimodal approaches primarily rely on feature fusion or cross-modal attention, which may not effectively capture hierarchical semantic inconsistencies across different levels of representation. To address this limitation, this paper proposes HCIG (Hierarchical Cross-modal Incongruity Graph Network), a novel framework that models cross-modal incongruity at token, phrase, and global levels using graph attention networks and adaptively integrates these representations through a learned hierarchical attention mechanism. As a complementary architecture, we also introduce GCCN (Graph-based Cross-modal Contradiction Network), which performs graph-based reasoning using contradiction-aware pooling for efficient multimodal interaction learning. The proposed models are evaluated on the MMSD sarcasm benchmark and the MultiBully cyberbullying dataset, together with comprehensive ablation studies and cross-task transfer experiments. Experimental results demonstrate that HCIG achieves the best performance on MMSD with 85.74\% accuracy and 85.29\% macro-F1, while GCCN attains the highest macro-F1 (68.66\%) on MultiBully and HCIG achieves the highest accuracy (69.62\%) and bullying-class F1 (74.90\%). The findings demonstrate that hierarchical multi-granularity incongruity modeling provides more effective multimodal reasoning than conventional fusion strategies, offering a robust framework for sarcasm and cyberbullying detection in social media. 
\end{abstract}

\section{Introduction}
Sarcasm and cyberbullying are both pragmatic phenomena that can be signaled as much by what a message shows as by what it says. A caption reading ``what a beautiful day'' paired with an image of a flooded street is sarcastic precisely because of the mismatch between text and image; a comment that reads as neutral in isolation can become bullying when paired with a mocking or humiliating image. Text-only classifiers, including large pretrained language models \cite{devlin2019bert}, systematically miss this class of example because the incongruity is only observable across modalities. This motivates multimodal approaches that explicitly model the \emph{relationship}, rather than only the concatenation, between textual and visual signals.

\begin{figure}[H]
\centering
\includegraphics[width=\textwidth]{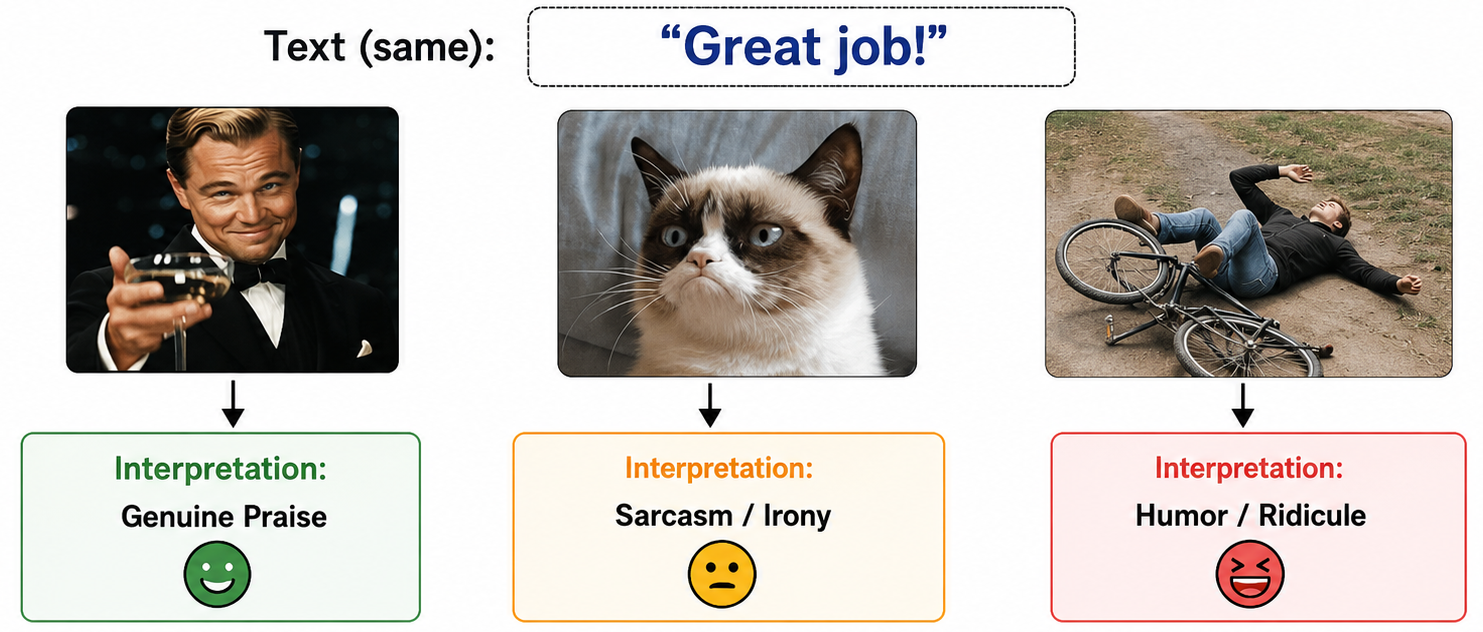}
\caption{Representative examples from a multimodal cyberbullying dataset highlighting the role of image–text interaction in distinguishing cyberbullying from non-bullying content.}
\label{fig:intro}
\end{figure}

Multimodal sarcasm detection has been studied through the MMSD benchmark of image--text pairs collected from social media \cite{cai2019multimodal,schifanella2016detecting}, with methods spanning hierarchical fusion \cite{cai2019multimodal}, cross-modality contrast modeling \cite{xu2020reasoning}, intra-/inter-modality incongruity modeling \cite{pan2020modeling}, and graph-based cross-modal reasoning \cite{liang2021interactive,liang2022crossmodal} intended to explicitly capture incongruity rather than relying on fusion alone. A revised, de-biased version of the benchmark, MMSD2.0, was later released to remove spurious textual cues \cite{qin2023mmsd2}. Cyberbullying detection has a longer history as a predominantly unimodal text classification problem \cite{kim2021humancentered,balakrisnan2023cyberbullying}; multimodal cyberbullying datasets that pair short text with an associated image, such as MultiBully \cite{maity2022multitask}, are comparatively recent, drawing on the broader multimodal hate-speech and meme-analysis literature established by benchmarks such as the Hateful Memes Challenge \cite{kiela2020hateful} and related multimodal hate-speech studies \cite{gomez2020exploring,lee2021disentangling,yang2022multimodal,hossain2024deciphering,das2020detecting,lippe2020multimodal,mozafari2020hate} as shown in Figure ~\ref{fig:intro}. It remains an open question whether incongruity-oriented architectures designed for sarcasm generalize to bullying detection, where the relevant signal (mockery, threat, or humiliation) need not involve semantic contradiction between modalities in the same way sarcasm does.

This paper addresses that gap with three contributions. First, and primarily, we propose \textbf{HCIG}, a hierarchical cross-modal incongruity graph network that computes incongruity independently at token, phrase, and global granularity and combines the three via a learned softmax attention gate, allowing the model to adaptively emphasize whichever granularity is most informative for a given example. Second, we implement \textbf{GCCN}, a structurally simpler graph-based architecture that represents text tokens and image patches as nodes in a similarity-thresholded graph reasoned over with a graph attention network (GATv2 \cite{brody2022attentive}) and augmented with a single learned contradiction-pooling score, which serves both as an independent point of comparison and as a design precursor that motivates HCIG's hierarchical extension. Third, we evaluate both architectures, together with unimodal and late-fusion baselines, on both MMSD (sarcasm) and MultiBully (cyberbullying), including in-domain training, MMSD-to-MultiBully fine-tuning, a bidirectional cross-task direct-transfer stress test, and component-level ablation studies for each architecture.

\section{Related Work}

\subsection{Multimodal sarcasm detection}
Schifanella et al.\cite{schifanella2016detecting} first framed multimodal sarcasm detection as a joint text--image classification problem. Cai et al.\cite{cai2019multimodal} introduced the MMSD benchmark used in this paper and a hierarchical fusion baseline; Qin et al. \cite{qin2023mmsd2} later identified and corrected spurious textual cues and mis-annotations in the original release, producing MMSD2.0. Subsequent work has increasingly modeled the text--image relationship explicitly rather than fusing modalities directly: Xu et al.\cite{xu2020reasoning} proposed a decomposition-and-relation network to capture cross-modality contrast and semantic association; Pan et al. \cite{pan2020modeling} modeled intra- and inter-modality incongruity with transformer-based attention; and Liang et al.\ introduced in-modal and cross-modal graph representations reasoned over with graph convolutional and graph attention mechanisms in two related studies \cite{liang2021interactive,liang2022crossmodal}, the line of work most directly related to the GCCN and HCIG architectures compared here. A separate line of work, MUStARD, extended multimodal sarcasm detection to video and text modalities drawn from television dialogue rather than static image--text social media posts \cite{castro2019towards}.

\subsection{Multimodal cyberbullying and hate-speech detection}
Cyberbullying detection has traditionally been treated as a unimodal text classification problem, with a substantial body of work on lexical, syntactic, and pretrained-language-model features \cite{kim2021humancentered,balakrisnan2023cyberbullying,allwaibed2025cyberbullying}. Multimodal treatments are more recent. The Hateful Memes Challenge established that unimodal models systematically fail on memes constructed so that hatefulness emerges only from the text--image combination \cite{kiela2020hateful}, motivating a growing body of multimodal hate-speech architectures \cite{gomez2020exploring,lee2021disentangling,yang2022multimodal,hossain2024deciphering,das2020detecting,lippe2020multimodal,mozafari2020hate}. Maity et al.\ extended this direction to cyberbullying specifically with MultiBully, a code-mixed, multi-label (bully, sentiment, emotion, sarcasm) meme dataset collected from Twitter and Reddit, which we use as the cyberbullying benchmark in this paper \cite{maity2022multitask}. To our knowledge, no prior work has directly compared graph- and hierarchy-based cross-modal incongruity architectures originally developed for sarcasm detection on this dataset, nor examined whether sarcasm-trained incongruity representations transfer to bullying detection.

\subsection{Multimodal sentiment, emotion, and meme analysis}
Beyond sarcasm and bullying specifically, a broader literature on multimodal sentiment and emotion analysis informs the fusion and attention design choices used here. Recent surveys taxonomize fusion strategies, learning paradigms, and attention mechanisms across the field \cite{verma2025navigating}, and graph convolutional approaches have been proposed for speaker-aware multimodal emotion recognition \cite{verma2026samer} and for conversational emotion recognition more broadly \cite{hu2021mmgcn}, motivating the use of graph attention as a cross-modal reasoning mechanism in this work. Attention-based frameworks have also been proposed specifically for multimodal sentiment analysis in memes \cite{verma2025mham}, the same content format used in MultiBully. Graph neural networks more generally have been surveyed extensively \cite{wu2021comprehensive}, and their foundational variants \cite{kipf2017semisupervised,velickovic2018graph,brody2022attentive} underpin the graph-reasoning modules in both GCCN and HCIG.

\subsection{Architectural components}
GCCN and HCIG both build on Transformer-based encoders \cite{vaswani2017attention}: RoBERTa \cite{liu2019roberta} for text and Vision Transformer \cite{dosovitskiy2021image} for images, trained with the AdamW optimizer \cite{loshchilov2019decoupled}, a decoupled-weight-decay variant of Adam \cite{kingma2015adam}. Both use graph attention networks for cross-modal reasoning, specifically GATv2 \cite{brody2022attentive}, which was proposed to correct a static-attention limitation in the original graph attention formulation \cite{velickovic2018graph}; the related graph convolutional network formulation \cite{kipf2017semisupervised} is noted here as an alternative, non-attentive graph reasoning mechanism not used in this work. Where relevant we also note CLIP-style joint vision--language pretraining \cite{radford2021learning} and ViLBERT-style two-stream co-attentional pretraining \cite{lu2019vilbert} as alternative encoder strategies not used in this study, and Grad-CAM \cite{selvaraju2017gradcam} as the visualization technique underlying part of the exploratory analysis in Section~\ref{sec:results}. Beyond the graph-attention formulation adopted here, entropic optimal transport \cite{cuturi2013sinkhorn} and Dirichlet-based evidential uncertainty estimation \cite{sensoy2018evidential} have both been proposed as alternative mechanisms for cross-modal alignment and confidence-aware prediction, respectively; neither is used by GCCN or HCIG in the present study, but both remain relevant directions for future extensions of this work.

\subsection{Summary comparison}
Table~\ref{tab:relwork} summarizes representative prior work by task, modality, core technique, and dataset, situating HCIG and GCCN relative to the existing literature.

\begin{table}[H]
\centering
\caption{Comparison of representative multimodal sarcasm and cyberbullying detection methods.}
\label{tab:relwork}
\small
\begin{tabular}{p{2.6cm}p{1.6cm}p{2.9cm}p{4.3cm}p{2.3cm}}
\toprule
\textbf{Study} & \textbf{Year} & \textbf{Task} & \textbf{Core Technique} & \textbf{Dataset(s)} \\
\midrule
Schifanella et al.\ \cite{schifanella2016detecting} & 2016 & Sarcasm & Hand-crafted + CNN visual features & Twitter/Instagram \\
Cai et al.\ \cite{cai2019multimodal} & 2019 & Sarcasm & Hierarchical fusion (text/image/attribute) & MMSD \\
Xu et al.\ \cite{xu2020reasoning} & 2020 & Sarcasm & Decomposition and relation network & MMSD \\
Pan et al.\ \cite{pan2020modeling} & 2020 & Sarcasm & Intra-/inter-modality transformer attention & MMSD \\
Liang et al.\ \cite{liang2021interactive} & 2021 & Sarcasm & In-modal + cross-modal graphs & MMSD \\
Liang et al.\ \cite{liang2022crossmodal} & 2022 & Sarcasm & Cross-modal graph convolutional network & MMSD \\
Qin et al.\ \cite{qin2023mmsd2} & 2023 & Sarcasm & Debiased benchmark + baseline re-evaluation & MMSD2.0 \\
Kiela et al.\ \cite{kiela2020hateful} & 2020 & Hate speech & Multimodal meme benchmark + baselines & Hateful Memes \\
Maity et al.\ \cite{maity2022multitask} & 2022 & Cyberbullying & Multitask sentiment/emotion/sarcasm-aware network & MultiBully \\
\textbf{This work (GCCN)} & -- & Sarcasm + Bullying & Similarity-graph + GATv2 + contradiction pooling & MMSD, MultiBully \\
\textbf{This work (HCIG)} & -- & Sarcasm + Bullying & Hierarchical token/phrase/global GATv2 + attention gate & MMSD, MultiBully \\
\bottomrule
\end{tabular}
\end{table}

\section{Proposed Methodology}
\label{sec:method}

\begin{figure}[H]
\centering
\includegraphics[width=\textwidth]{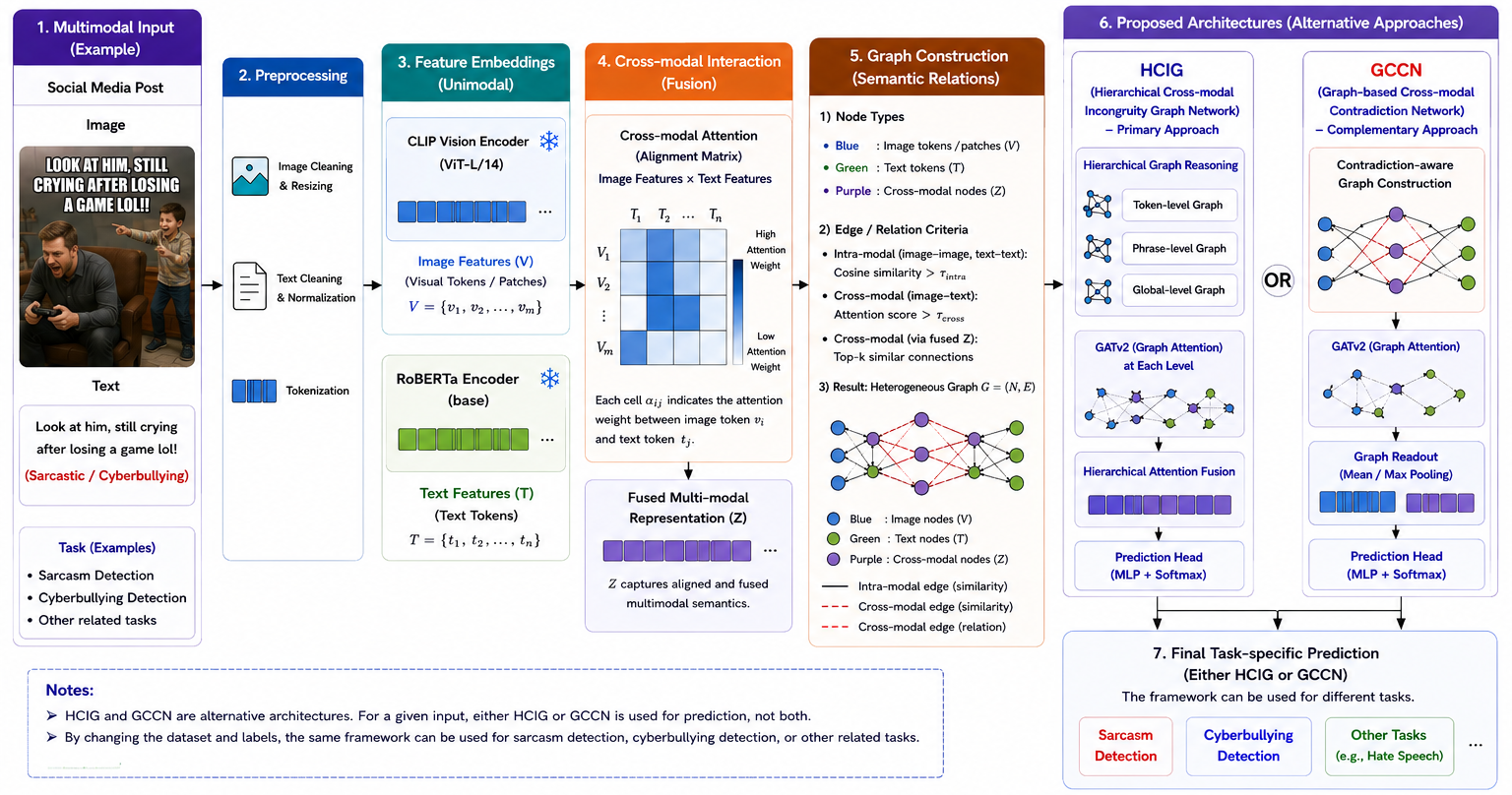}
\caption{Conceptual overview of the shared multimodal input, preprocessing, encoder, cross-modal interaction, and graph-construction stages, followed by the alternative HCIG and GCCN prediction architectures.}
\label{fig:architecture_overview}
\end{figure}

Figure~\ref{fig:architecture_overview} provides a conceptual overview of the shared multimodal processing pipeline and positions HCIG and GCCN as alternative task-specific architectures. The detailed computational flows of GCCN and HCIG are shown in Figures~\ref{fig:gccn_architecture} and~\ref{fig:hcig_architecture}, respectively; the exact operations implemented by the models are defined mathematically in Sections~\ref{sec:gccn} and~\ref{sec:hcig}.

\subsection{Problem Formulation and Notation}
\label{sec:formulation}
We formulate multimodal sarcasm and cyberbullying detection as binary classification over paired text--image inputs. Let $\mathcal{D} = \{(t_i, v_i, y_i)\}_{i=1}^{N}$ denote a dataset of $N$ text--image pairs, where $t_i$ is a social-media post's text, $v_i$ is its associated image, and $y_i \in \{0,1\}$ is a binary label (sarcastic/non-sarcastic for MMSD, bully/non-bully for MultiBully). The text $t_i$ is tokenized into a sequence of $n$ subword tokens $\{w_1,\dots,w_n\}$ and encoded by a text Transformer to produce contextual token representations $H^{t}_i \in \mathbb{R}^{n \times d}$ and a pooled sentence representation $h^{t}_{\text{CLS},i} \in \mathbb{R}^{d}$. The image $v_i$ is divided into $m$ fixed-size patches and encoded by a vision Transformer to produce patch representations $H^{v}_i \in \mathbb{R}^{m \times d}$ and a pooled image representation $h^{v}_{\text{CLS},i} \in \mathbb{R}^{d}$, with $d$ the shared hidden dimensionality of both encoders.

The goal is to learn a function $f_\theta : (t_i, v_i) \mapsto \hat{y}_i$ that predicts $\hat{y}_i \approx y_i$ while explicitly modeling a cross-modal \emph{incongruity} signal $\delta(H^t_i, H^v_i)$ that captures the degree of semantic mismatch between the text and image, rather than relying only on a fused joint representation $[h^t_{\text{CLS},i}; h^v_{\text{CLS},i}]$. We denote by $z_i$ the learned incongruity-aware representation produced by an architecture's cross-modal reasoning module, which is combined with the pooled unimodal representations and passed to a classification head $g_\theta(\cdot)$:
\begin{equation}
\hat{y}_i = g_\theta\big([\,z_i \,;\, h^{t}_{\text{CLS},i} \,;\, h^{v}_{\text{CLS},i}\,]\big)
\label{eq:general_form}
\end{equation}
GCCN and HCIG, described below, differ in how $z_i$ is computed and, in HCIG's case, in whether $h^t_{\text{CLS},i}$ and $h^v_{\text{CLS},i}$ are concatenated again at the classification head or are already subsumed into $z_i$ (Section~\ref{sec:hcig}).

\subsection{Backbone Encoders}
\label{sec:backbone}
Both architectures share a common backbone: text is encoded with RoBERTa-base \cite{liu2019roberta} and images with a ViT-Base/16 Vision Transformer \cite{dosovitskiy2021image}, both fine-tuned end-to-end, following the general Transformer encoder design of Vaswani et al.\ \cite{vaswani2017attention}. Text sequences are tokenized to a maximum length of 80 subwords; images are resized to $224\times224$.

Image features require an explicit alignment step because ViT backbones expose a class token concatenated with patch tokens, which must be separated before cross-modal reasoning. Writing $\phi_{\text{ViT}}(v_i)\in\mathbb{R}^{(m+1)\times d_v}$ for the raw ViT feature sequence and $W_v\in\mathbb{R}^{d\times d_v}$ for a learned projection to the shared hidden size,
\begin{equation}
\tilde H^{v}_i = W_v\, \phi_{\text{ViT}}(v_i) \in \mathbb{R}^{(m+1)\times d}, \qquad
h^{v}_{\text{CLS},i} = \tilde H^{v}_i[0], \qquad H^{v}_i = \tilde H^{v}_i[1:],
\label{eq:vit_split}
\end{equation}
so that $h^v_{\text{CLS},i}$ and the $m$ patch representations $H^v_i$ used downstream are disjoint; backbones without a class token instead pool a single pseudo-patch and reuse it as $h^v_{\text{CLS},i}$.

On the text side, the RoBERTa output sequence contains a beginning-of-sequence (BOS) token, an end-of-sequence (EOS) token, and right-padding, none of which should participate in token-level cross-modal reasoning. Let $a_{i,j}\in\{0,1\}$ be the attention mask returned by the tokenizer and $\ell_i=\sum_j a_{i,j}$ the true sequence length (including BOS/EOS). The BOS position is used as $h^t_{\text{CLS},i}$; the remaining $n{-}1$ positions form $H^t_i$, with a token validity mask
\begin{equation}
\mathcal{M}^{t}_i[j] = \mathbb{1}\big[\,j < \ell_i - 1\,\big], \qquad j = 1,\dots,n-1,
\label{eq:token_mask}
\end{equation}
that excludes BOS, EOS, and padding, so that only genuine subword tokens can enter the contradiction- and incongruity-scoring modules described below. We write $\mathcal{T}_i=\{j : \mathcal{M}^t_i[j]=1\}$ for the resulting set of valid token indices.

\subsection{Shared Cross-Modal Graph Construction and Reasoning}
\label{sec:graphprim}
GCCN and HCIG both reason over text--image node sets using the same graph-construction-and-propagation primitive, applied at different granularities; we describe it once here and reuse it by reference in Sections~\ref{sec:gccn}--\ref{sec:hcig}. Given a set of text-side nodes $\{x_1,\dots,x_p\}$ (individual tokens, pooled phrases, or a single CLS vector, depending on level) and image-side nodes $\{x_{p+1},\dots,x_{p+q}\}$ (patches or a single CLS vector), each node is first tagged with a learned modality embedding $e_{\text{mod}}:\{0,1\}\to\mathbb{R}^d$,
\begin{equation}
\hat x_u = x_u + e_{\text{mod}}(\mathbb{1}[u \text{ is an image node}]), \qquad u = 1,\dots,p+q,
\label{eq:node_emb}
\end{equation}
so the subsequent graph attention can distinguish node modality independently of content. A weighted graph is then built by thresholding pairwise cosine similarity, with self-loops inserted explicitly rather than learned:
\begin{equation}
s_{uw} = \frac{\hat x_u \cdot \hat x_w}{\lVert \hat x_u\rVert\,\lVert \hat x_w\rVert + \epsilon}, \qquad
\mathcal{A}_{uw} = \mathbb{1}[\,s_{uw} > \tau\,] \ \lor\ \mathbb{1}[u=w],
\label{eq:adjacency}
\end{equation}
giving a sparse edge set $\mathcal{E}=\{(u,w) : \mathcal{A}_{uw}=1\}$ with each retained edge carrying its similarity as an edge attribute, $e_{uw}=s_{uw}$. A two-layer GATv2 network \cite{brody2022attentive} propagates over $(\{\hat x_u\}, \mathcal{E}, \{e_{uw}\})$; for a single attention head, the update at node $u$ over neighborhood $\mathcal{N}(u)$ is
\begin{equation}
\alpha_{uw} = \frac{\exp\big(a^\top \mathrm{LeakyReLU}(W_1 \hat x_u + W_2 \hat x_w + W_e\, e_{uw})\big)}
{\sum_{w'\in\mathcal{N}(u)} \exp\big(a^\top \mathrm{LeakyReLU}(W_1 \hat x_u + W_2 \hat x_{w'} + W_e\, e_{uw'})\big)},
\label{eq:gat_attn}
\end{equation}
\begin{equation}
X^{(1)} = \mathrm{GELU}\Big(\big[\textstyle\sum_{w\in\mathcal{N}(u)} \alpha^{(h)}_{uw} W_2^{(h)} \hat x_w\big]_{h=1}^{H}\Big), \qquad
X^{(2)} = \mathrm{GATv2}\big(X^{(1)}, \mathcal{E}, \{e_{uw}\}\big),
\label{eq:gat_layer}
\end{equation}
where the first layer concatenates $H$ attention heads (Eq.~\ref{eq:gat_attn} instantiated per head) and the second layer uses a single head to return node states in $\mathbb{R}^d$. We denote this two-stage operator compactly as $\mathrm{GraphReason}(\{x_u\}, \mathcal{E})$ in the remainder of this section.

\subsection{GCCN: Graph-based Cross-modal Contradiction Network}
\label{sec:gccn}
GCCN instantiates a single application of $\mathrm{GraphReason}$ over the full set of valid text tokens $\{h^t_j : j\in\mathcal{T}_i\}$ and all $m{+}1$ image nodes (patches plus CLS), following Eqs.~\ref{eq:node_emb}--\ref{eq:gat_layer} with threshold $\tau=0.3$, and mean-pools the resulting node states into a single graph-level embedding:
\begin{equation}
z^{\text{graph}}_i = \frac{1}{|\mathcal{T}_i| + m + 1} \sum_{u} X^{(2)}_u 
\label{eq:graph_pool}
\end{equation}
In parallel, a dedicated contradiction-pooling module scores every valid token--patch pair directly, independently of the graph pathway. For token $j\in\mathcal{T}_i$ and patch $l\in\{1,\dots,m\}$,
\begin{equation}
s_{jl} = \frac{h^t_j \cdot h^v_l}{\lVert h^t_j\rVert\,\lVert h^v_l\rVert + \epsilon}, \qquad d_{jl} = \mathrm{clip}(1 - s_{jl},\, 0,\, 2),
\label{eq:gccn_sim}
\end{equation}
and a small learned gate, conditioned on similarity, distance, and squared similarity, controls how much each pair contributes to the contradiction score:
\begin{equation}
C_{jl} = \sigma\big(\mathrm{MLP}_c([\,s_{jl}\,;\,d_{jl}\,;\,s_{jl}^2\,])\big)\cdot d_{jl}
\label{eq:gccn_gate}
\end{equation}
A single scalar contradiction score per example is obtained by masked top-$k$ pooling ($k{=}10$) over the valid entries of the contradiction matrix $C_i=\{C_{jl}\}_{j\in\mathcal{T}_i,\,l\le m}$,
\begin{equation}
c_i = \frac{1}{\min(k,|\mathrm{valid}(C_i)|)} \sum_{(j,l)\,\in\,\mathrm{TopK}_k(\mathrm{valid}(C_i))} C_{jl},
\label{eq:gccn_topk}
\end{equation}
so that $c_i$ reflects the most strongly contradicted token--patch pairs for the example rather than an average dominated by well-aligned pairs. GCCN instantiates Eq.~\ref{eq:general_form} with
\begin{equation}
z_i = [\,z^{\text{graph}}_i \,;\, c_i\,]
\label{eq:gccn_z}
\end{equation}

\begin{figure}[H]
\centering
\includegraphics[width=\textwidth]{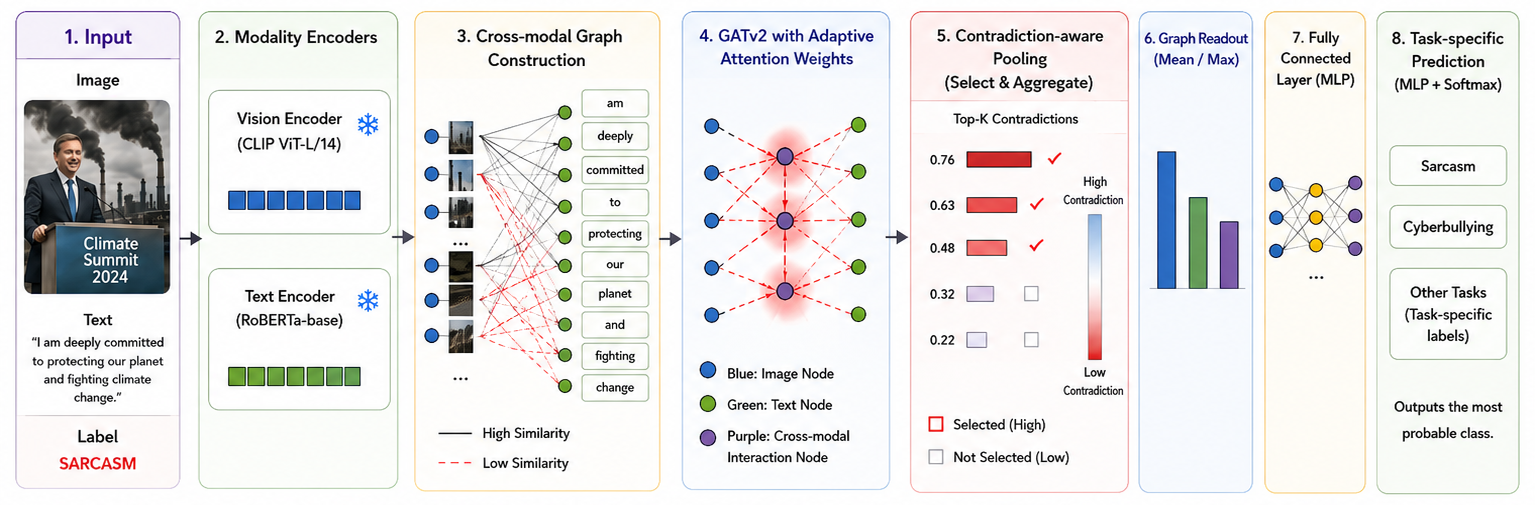}
\caption{Conceptual architecture of GCCN, showing multimodal encoding, cross-modal graph construction, adaptive GATv2 reasoning, contradiction-aware pooling, graph readout, and task-specific classification.}
\label{fig:gccn_architecture}
\end{figure}

The end-to-end GCCN processing sequence, including modality encoding, graph construction, adaptive GATv2 reasoning, contradiction-aware pooling, graph readout, and task-specific prediction, is illustrated conceptually in Figure~\ref{fig:gccn_architecture}.

\subsection{HCIG: Hierarchical Cross-modal Incongruity Graph Network }
\label{sec:hcig}
HCIG, the primary architecture proposed in this work, computes incongruity at three levels of granularity rather than a single graph pass, motivated by the observation that cross-modal incongruity in sarcasm and bullying content can manifest as an individual word--object mismatch (token level), a short phrase-level mismatch (phrase level), or a single global tone mismatch between the overall message and image (global level). Token and phrase levels share the same three-stage procedure --- graph reasoning, incongruity-guided alignment, residual fusion --- applied to different text granularities against the same set of image patches; the global level instead operates directly on the pooled [CLS] representations.

\textbf{Phrase construction.} Before phrase-level reasoning, valid text tokens are mean-pooled into non-overlapping, mask-aware groups of three:
\begin{equation}
h^{t,\text{phr}}_{i,g} = \frac{\sum_{r=1}^{3} \mu_{i,g,r}\, h^t_{i,\,3(g-1)+r}}{\sum_{r=1}^{3}\mu_{i,g,r} + \epsilon}, \qquad
\mu_{i,g} = \bigvee_{r=1}^{3}\mu_{i,g,r},
\label{eq:phrase_pool}
\end{equation}
for $g=1,\dots,\lceil n/3\rceil$, where $\mu_{i,g,r}\in\{0,1\}$ marks whether the $r$-th token of group $g$ is a valid (non-padding) token per Eq.~\ref{eq:token_mask}; groups with no valid tokens are masked out entirely by $\mu_{i,g}=0$.

\textbf{Level graph reasoning.} At each level $\text{lvl}\in\{\text{tok},\text{phr}\}$, the level's text units (tokens or phrases) and the $m$ image patches are passed through $\mathrm{GraphReason}$ (Eqs.~\ref{eq:node_emb}--\ref{eq:gat_layer}) using level-specific GATv2 weights, producing graph-updated text units $\{\tilde h^{\text{lvl}}_u\}$ and graph-updated patches $\{\tilde h^{v}_l\}$.

\textbf{Incongruity-guided cross-modal alignment.} For each graph-updated text unit $u$ at level $\text{lvl}$ and each graph-updated patch $l$,
\begin{equation}
s_{ul} = \cos(\tilde h^{\text{lvl}}_u, \tilde h^v_l), \qquad d_{ul} = \mathrm{clip}(1-s_{ul}, 0, 2), \qquad
\iota_{ul} = \sigma\big(\mathrm{MLP}_\iota([\,s_{ul}\,;\,d_{ul}\,;\,s_{ul}^2\,])\big)\cdot d_{ul},
\label{eq:hcig_sim}
\end{equation}
using a single $\mathrm{MLP}_\iota$ shared across the token and phrase levels. Rather than aligning each text unit to its most \emph{similar} patch, HCIG deliberately attends toward the most \emph{divergent} patches, since it is precisely the mismatched regions of the image that carry incongruity signal:
\begin{equation}
w_{ul} = \frac{\exp(d_{ul})}{\sum_{l'} \exp(d_{ul'})}, \qquad \tilde a_u = \sum_{l} w_{ul}\, \tilde h^v_l
\label{eq:hcig_align}
\end{equation}
Each text unit is then updated with a residual fusion of itself and its incongruity-aligned image counterpart,
\begin{equation}
r_u = \mathrm{LN}\Big(\mathrm{MLP}_r\big([\,\tilde h^{\text{lvl}}_u\,;\,\tilde a_u\,;\,|\tilde h^{\text{lvl}}_u - \tilde a_u|\,;\,\tilde h^{\text{lvl}}_u \odot \tilde a_u\,]\big)\Big),
\label{eq:hcig_resid}
\end{equation}
and the level representation is obtained by pooling residual states, weighted by each unit's peak incongruity across patches, $\omega_u = \max_l \iota_{ul}$:
\begin{equation}
z^{\text{lvl}}_i = \frac{\sum_u \omega_u\, r_u}{\sum_u \omega_u + \epsilon}, \qquad
\iota^{\text{lvl}}_i = \frac{1}{|\mathcal{U}_i|\cdot m}\sum_{u\in\mathcal{U}_i}\sum_{l} \iota_{ul},
\label{eq:hcig_levelpool}
\end{equation}
where $\mathcal{U}_i$ is the set of valid units at that level ($\mathcal{T}_i$ for tokens, valid phrase groups for phrases) and $\iota^{\text{lvl}}_i$ is a scalar level-incongruity score used both for reporting and for the fusion gate below.

\textbf{Global level.} The pooled [CLS] representations are compared directly, without graph reasoning, via the same pairwise-interaction pattern used at the token and phrase levels:
\begin{equation}
\begin{aligned}
z^{\text{glob}}_i
&= \mathrm{LN}\Big(\mathrm{MLP}_g\big([
    \,h^t_{\text{CLS},i}\,;\,h^v_{\text{CLS},i}\,; \\
&\hspace{4.7em}
    \,|h^t_{\text{CLS},i}-h^v_{\text{CLS},i}|\,;\,
    h^t_{\text{CLS},i}\odot h^v_{\text{CLS},i}\,]
\big)\Big), \\
\iota^{\text{glob}}_i
&= 1 - \cos(h^t_{\text{CLS},i}, h^v_{\text{CLS},i})
\end{aligned}
\label{eq:hcig_global}
\end{equation}

\textbf{Level-attention fusion.} The three level representations and their scalar incongruity scores are combined with a learned softmax attention gate rather than fixed or uniform weighting:
\begin{equation}
\boldsymbol{\alpha}_i = \mathrm{softmax}\Big(\mathrm{MLP}_\alpha\big([\,z^{\text{tok}}_i\,;\,z^{\text{phr}}_i\,;\,z^{\text{glob}}_i\,;\,\iota^{\text{tok}}_i\,;\,\iota^{\text{phr}}_i\,;\,\iota^{\text{glob}}_i\,]\big)\Big) \in \Delta^2,
\label{eq:hcig_gate}
\end{equation}
so the gate can condition its weighting on how incongruent each level already appears to be, not only on the content of each level's representation. The gated level representations are then \emph{concatenated} --- preserving each level's identity within the fused vector rather than collapsing them into a single $d$-dimensional average --- and passed to the classifier:
\begin{equation}
z_i = \big[\,\alpha^{\text{tok}}_i\, z^{\text{tok}}_i \,;\, \alpha^{\text{phr}}_i\, z^{\text{phr}}_i \,;\, \alpha^{\text{glob}}_i\, z^{\text{glob}}_i\,\big] \in \mathbb{R}^{3d}, \qquad
\hat{y}_i = g_\theta(z_i),
\label{eq:hcig_fused}
\end{equation}
which departs from the general fusion form of Eq.~\ref{eq:general_form}: because $z^{\text{glob}}_i$ (Eq.~\ref{eq:hcig_global}) already integrates $h^t_{\text{CLS},i}$ and $h^v_{\text{CLS},i}$ through their pairwise interaction, HCIG's classifier operates directly on $z_i$ rather than re-concatenating the raw CLS vectors, avoiding redundant duplication of the same information. A single scalar incongruity score is additionally defined as the attention-weighted combination of the three level scores,
\begin{equation}
\iota_i = \alpha^{\text{tok}}_i\, \iota^{\text{tok}}_i + \alpha^{\text{phr}}_i\, \iota^{\text{phr}}_i + \alpha^{\text{glob}}_i\, \iota^{\text{glob}}_i
\label{eq:hcig_inconscore}
\end{equation}
All GATv2 layers across the three levels are genuinely active in the forward pass, correcting an implementation issue present in an earlier development version of this architecture in which GAT layers were declared but not used; this is noted explicitly because it is directly relevant to interpreting the corresponding ablation results in Section~\ref{sec:ablation}.

\begin{figure}[H]
\centering
\includegraphics[width=\textwidth]{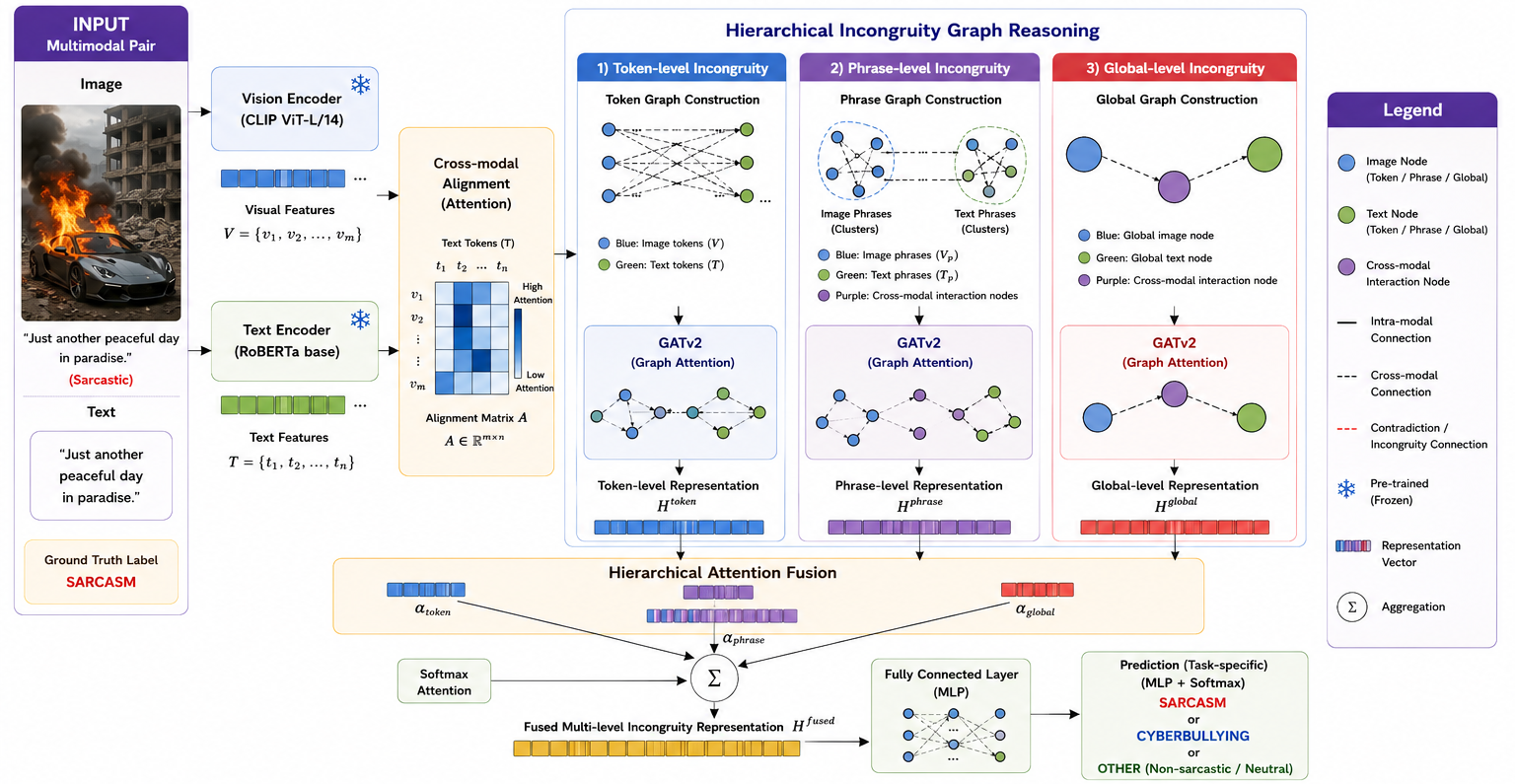}
\caption{Conceptual architecture of HCIG, showing token-, phrase-, and global-level incongruity reasoning, GATv2-based graph processing, hierarchical attention fusion, and task-specific classification.}
\label{fig:hcig_architecture}
\end{figure}

Figure~\ref{fig:hcig_architecture} summarizes the hierarchical HCIG workflow, from multimodal feature extraction and cross-modal alignment to token-, phrase-, and global-level graph reasoning, hierarchical attention fusion, and task-specific prediction.

\subsection{Training Objective}
Both architectures are trained end-to-end with a standard binary cross-entropy classification loss on $\hat{y}_i$ against $y_i$, using the AdamW optimizer \cite{loshchilov2019decoupled} with a linear warmup schedule. No auxiliary losses (e.g., contrastive, counterfactual, or uncertainty-regularization terms) are used for either GCCN or HCIG in this work; both rely solely on the classification objective operating on the incongruity-aware representation $z_i$ defined above.

\subsection{Algorithms}
\label{sec:algorithm}
Algorithm~\ref{alg:gccn} summarizes the GCCN forward pass (Section~\ref{sec:gccn}), and Algorithm~\ref{alg:hcig} summarizes the HCIG forward pass (Section~\ref{sec:hcig}), which is the main architectural contribution of this paper.

\begin{algorithm}[H]
\caption{\textit{GCCN Forward Pass}}
\label{alg:gccn}
\begin{algorithmic}[1]
\Require Text tokens $t$, image $v$
\Ensure Prediction $\hat{y}$, contradiction score $c$
\State $H^t, h^t_{\text{CLS}} \gets \text{RoBERTa}(t)$; \quad $H^v, h^v_{\text{CLS}} \gets \text{ViT}(v)$ \Comment{Eq.~\ref{eq:vit_split}}
\State $\mathcal{T} \gets \{j : \mathcal{M}^t[j]=1\}$ \Comment{Valid token mask, Eq.~\ref{eq:token_mask}}
\State $G \gets \text{BuildSimilarityGraph}(\{H^t_j\}_{j\in\mathcal{T}}, H^v \cup h^v_{\text{CLS}}, \tau)$ \Comment{Eqs.~\ref{eq:node_emb}--\ref{eq:adjacency}}
\State $X \gets \text{GraphReason}(G)$ \Comment{Eqs.~\ref{eq:gat_attn}--\ref{eq:gat_layer}}
\State $z^{\text{graph}} \gets \text{MeanPool}(X)$ \Comment{Eq.~\ref{eq:graph_pool}}
\State $C \gets \text{GatedContradiction}(\{H^t_j\}_{j\in\mathcal{T}}, H^v)$ \Comment{Eqs.~\ref{eq:gccn_sim}--\ref{eq:gccn_gate}}
\State $c \gets \text{MaskedTopK}(C, k{=}10)$ \Comment{Eq.~\ref{eq:gccn_topk}}
\State $z \gets [z^{\text{graph}}; c]$ \Comment{Eq.~\ref{eq:gccn_z}}
\State $\hat{y} \gets \text{Classifier}([z; h^t_{\text{CLS}}; h^v_{\text{CLS}}])$ \Comment{Eq.~\ref{eq:general_form}}
\State \Return $\hat{y}, c$
\end{algorithmic}
\end{algorithm}

\begin{algorithm}[H]
\caption{\textit{HCIG Forward Pass}}
\label{alg:hcig}
\begin{algorithmic}[1]
\Require Text tokens $t$, image $v$
\Ensure Prediction $\hat{y}$, level-attention weights $\boldsymbol{\alpha}$
\State $H^t, h^t_{\text{CLS}} \gets \text{RoBERTa}(t)$; \quad $H^v, h^v_{\text{CLS}} \gets \text{ViT}(v)$ \Comment{Eq.~\ref{eq:vit_split}}
\State $G_{\text{tok}} \gets \text{BuildSimilarityGraph}(H^t, H^v, \tau)$; \quad $\tilde H^t, \tilde H^v \gets \text{GraphReason}(G_{\text{tok}})$
\State $z^{\text{tok}}, \iota^{\text{tok}} \gets \text{IncongruityAlign}(\tilde H^t, \tilde H^v)$ \Comment{Eqs.~\ref{eq:hcig_sim}--\ref{eq:hcig_levelpool}}
\State $H^{t,\text{phr}} \gets \text{PhrasePool}(H^t, \text{size}{=}3)$ \Comment{Eq.~\ref{eq:phrase_pool}}
\State $G_{\text{phr}} \gets \text{BuildSimilarityGraph}(H^{t,\text{phr}}, H^v, \tau)$; \quad $\tilde H^{t,\text{phr}}, \tilde H^{v}_{\text{phr}} \gets \text{GraphReason}(G_{\text{phr}})$
\State $z^{\text{phr}}, \iota^{\text{phr}} \gets \text{IncongruityAlign}(\tilde H^{t,\text{phr}}, \tilde H^{v}_{\text{phr}})$ \Comment{Eqs.~\ref{eq:hcig_sim}--\ref{eq:hcig_levelpool}}
\State $z^{\text{glob}}, \iota^{\text{glob}} \gets \text{GlobalContradiction}(h^t_{\text{CLS}}, h^v_{\text{CLS}})$ \Comment{Eq.~\ref{eq:hcig_global}}
\State $\boldsymbol{\alpha} \gets \text{Softmax}(\text{MLP}([z^{\text{tok}}; z^{\text{phr}}; z^{\text{glob}}; \iota^{\text{tok}}; \iota^{\text{phr}}; \iota^{\text{glob}}]))$ \Comment{Level-attention gate, Eq.~\ref{eq:hcig_gate}}
\State $z \gets [\alpha^{\text{tok}} z^{\text{tok}}; \alpha^{\text{phr}} z^{\text{phr}}; \alpha^{\text{glob}} z^{\text{glob}}]$ \Comment{Eq.~\ref{eq:hcig_fused}}
\State $\hat{y} \gets \text{Classifier}(z)$ \Comment{Eq.~\ref{eq:hcig_fused}}
\State \Return $\hat{y}, \boldsymbol{\alpha}$
\end{algorithmic}
\end{algorithm}

\section{Implementation Details}
\label{sec:implementation}

\subsection{Dataset and Exploratory Analysis}
\label{sec:data}

\subsubsection{Multimodal Sarcasm Detection (MMSD)}
\begin{figure}[H]
\centering
\includegraphics[width=0.95\textwidth]{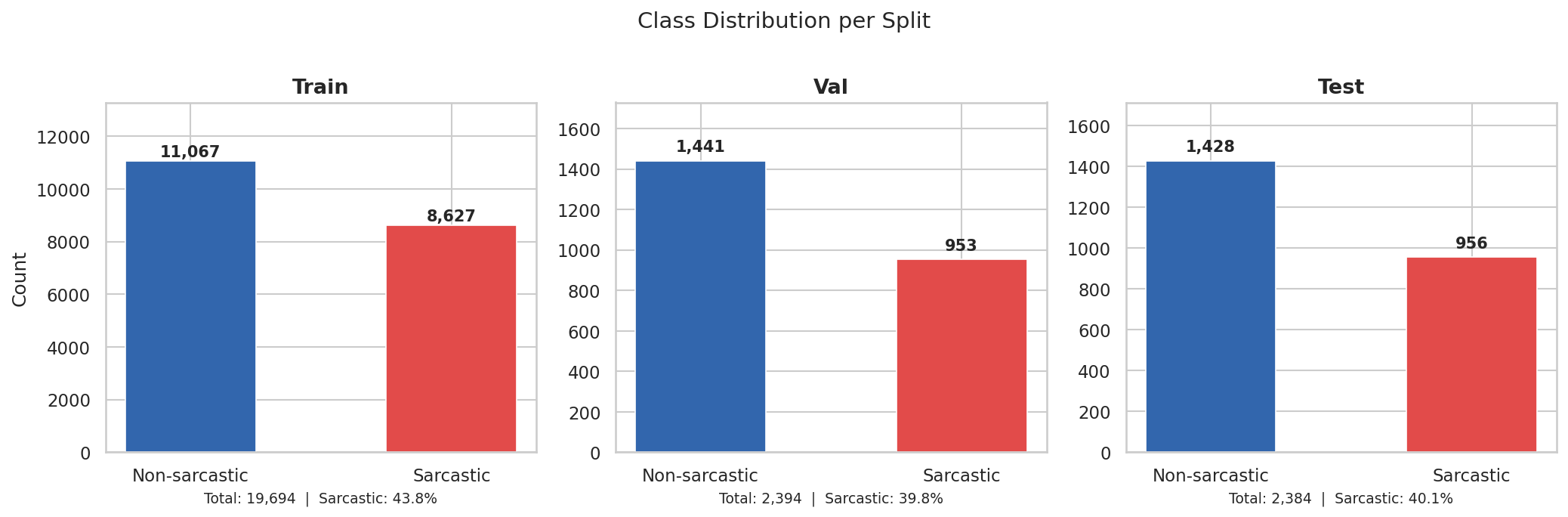}
\caption{MMSD class distribution by split (train/val/test), sarcastic vs.\ non-sarcastic.}
\label{fig:mmsd_dist}
\end{figure}

The raw MMSD release \cite{cai2019multimodal} comprises 33{,}696 text--image pairs labeled sarcastic or non-sarcastic. After removing rows with missing or corrupted image files, 24{,}472 pairs (72.6\% of the raw set) remained for experiments. The drop was not uniform across splits: 31.9\% of training rows were removed for missing images, versus 0\% of validation and test rows, because the validation and test partitions had already been curated in the source release. This asymmetry is reported explicitly because it means the effective training set is smaller, and its class balance and image-quality profile is not guaranteed to match the untouched validation/test partitions. Table~\ref{tab:dataset} and Figure~\ref{fig:mmsd_dist} summarize the cleaned splits and their class distributions.

\subsubsection{MultiBully}
\begin{figure}[H]
\centering
\includegraphics[width=0.95\textwidth]{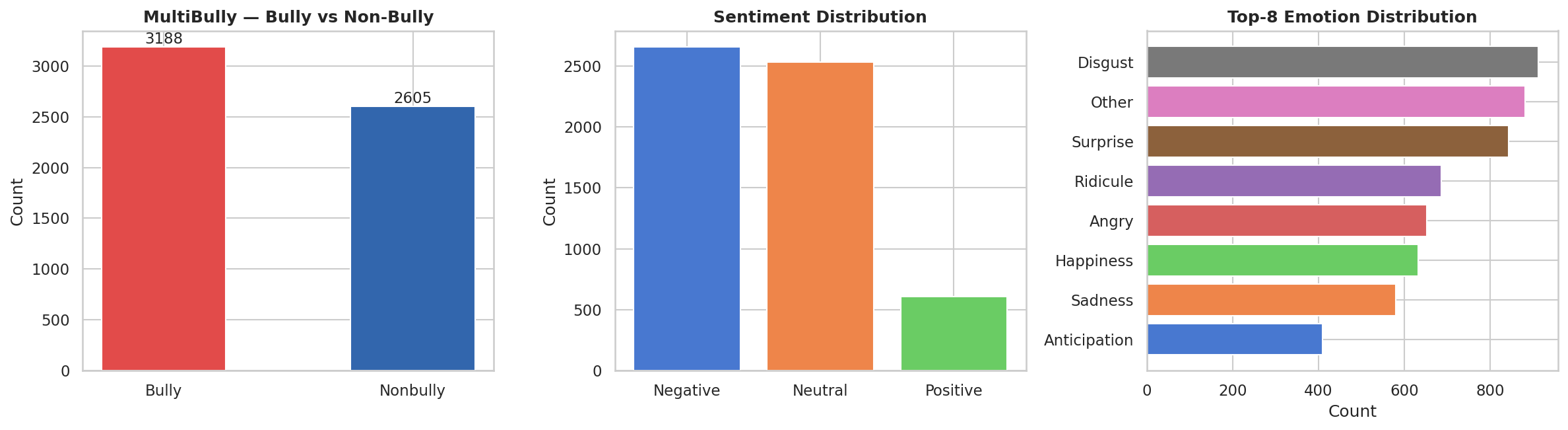}
\caption{Label and sentiment distribution in the cleaned MultiBully dataset.}
\label{fig:mb_dist}
\end{figure}

The MultiBully dataset \cite{maity2022multitask} pairs short social-media text with an associated image and a binary bully/non-bully label, collected in a code-mixed language setting from Twitter and Reddit. After removing 61 pairs (1.0\%) with missing or unreadable images, 5{,}793 pairs remained, comprising 3{,}188 bully and 2{,}605 non-bully examples (Table~\ref{tab:dataset}, Figure~\ref{fig:mb_dist}), a moderate class imbalance retained without resampling in all reported experiments.

\begin{table}[H]
\centering
\caption{ Statistics of the cleaned MMSD and MultiBully dataset. }
\label{tab:dataset}
\begin{tabular}{llrrr}
\toprule
Dataset & Split & Total pairs & Negative class & Positive class \\
\midrule
MMSD & train & 19{,}694 & 11{,}067 (Non-sarcastic) & 8{,}627 (Sarcastic) \\
MMSD & val   & 2{,}394  & 1{,}441 (Non-sarcastic)  & 953 (Sarcastic) \\
MMSD & test  & 2{,}384  & 1{,}428 (Non-sarcastic)  & 956 (Sarcastic) \\
MultiBully & all (cleaned) & 5{,}793 & 2{,}605 (Non-bully) & 3{,}188 (Bully) \\
\bottomrule
\end{tabular}
\end{table}

\subsection{Baselines}
Three baselines are reported: (i) \textbf{Text-BERT}, a BERT-base-uncased classifier \cite{devlin2019bert} using only the text modality; (ii) \textbf{Image-ResNet}, a fine-tuned ResNet-50 classifier \cite{he2016deep} using only the image modality; and (iii) \textbf{Late Fusion}, which concatenates pooled text and image representations from the shared RoBERTa/ViT backbone before a classification head, without any explicit incongruity modeling.

\subsection{Training Configuration and Experimental Setup}
Table~\ref{tab:trainconfig} summarizes the shared training configuration used across all models, baselines, ablations, and cross-task experiments unless the ablated component was itself the subject of the change. Three evaluation regimes are reported: (1) \textbf{in-domain MMSD}, where all five models (three baselines, GCCN, HCIG) are trained and evaluated on the MMSD train/validation/test splits described in Section~\ref{sec:data}; (2) \textbf{in-domain and fine-tuned MultiBully}, where the same five architectures are trained from randomly initialized task heads on MultiBully (``in-domain''), and GCCN/HCIG are additionally initialized from their complete MMSD-trained checkpoints before continued training on MultiBully (``fine-tuned''). In this fine-tuning regime, the existing two-class MMSD classification head is retained rather than reinitialized or replaced, and all loaded parameters are further optimized on MultiBully; and (3) a \textbf{bidirectional cross-task direct-transfer stress test}, where models trained only on MMSD are applied directly to the MultiBully test set with no target-task training, and vice versa. 

\begin{table}[H]
\centering
\caption{Experimental settings and training hyperparameters.}
\label{tab:trainconfig}
\small
\begin{tabular}{@{}p{0.28\linewidth}p{0.66\linewidth}@{}}
\toprule
\textbf{Setting} & \textbf{Value / Description} \\
\midrule
Optimizer & AdamW \cite{loshchilov2019decoupled}, linear warmup schedule \\
Batch size & 6 (constrained by combined RoBERTa, ViT, and graph-module memory footprint) \\
Max epochs & 10, with early stopping (patience $=3$) on validation macro-F1 \\
Text max length & 80 subword tokens \\
Image size & $224 \times 224$ \\
Graph similarity threshold  & 0.3 \\
Contradiction top-$k$ ($k$) & 10 \\
Phrase group size & 3 tokens \\
Evaluation regimes & In-domain MMSD; in-domain and fine-tuned MultiBully; bidirectional cross-task direct-transfer stress test \\
Reported metrics & Accuracy, Precision, Recall, F1, Macro-F1, ROC-AUC  \\
\bottomrule
\end{tabular}
\end{table}

\section{Results}
\label{sec:results}

\subsection{MMSD in-domain results}
\begin{figure}[H]
\centering
\includegraphics[width=\textwidth]{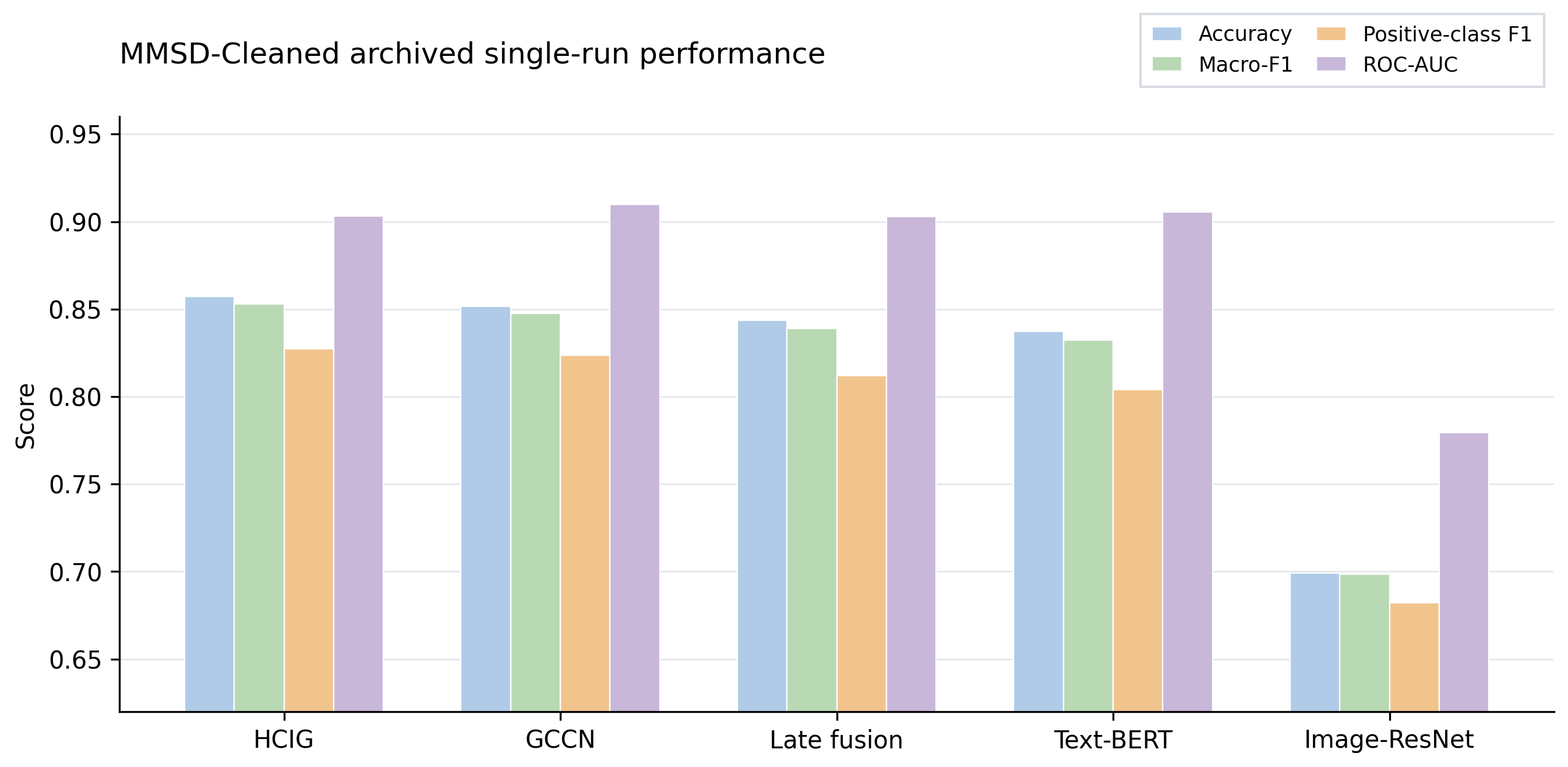}
\caption{Performance comparison of baseline and proposed models on MMSD.}
\label{fig:main_bar}
\end{figure}

\begin{figure}[H]
\centering
\includegraphics[width=0.95\textwidth]{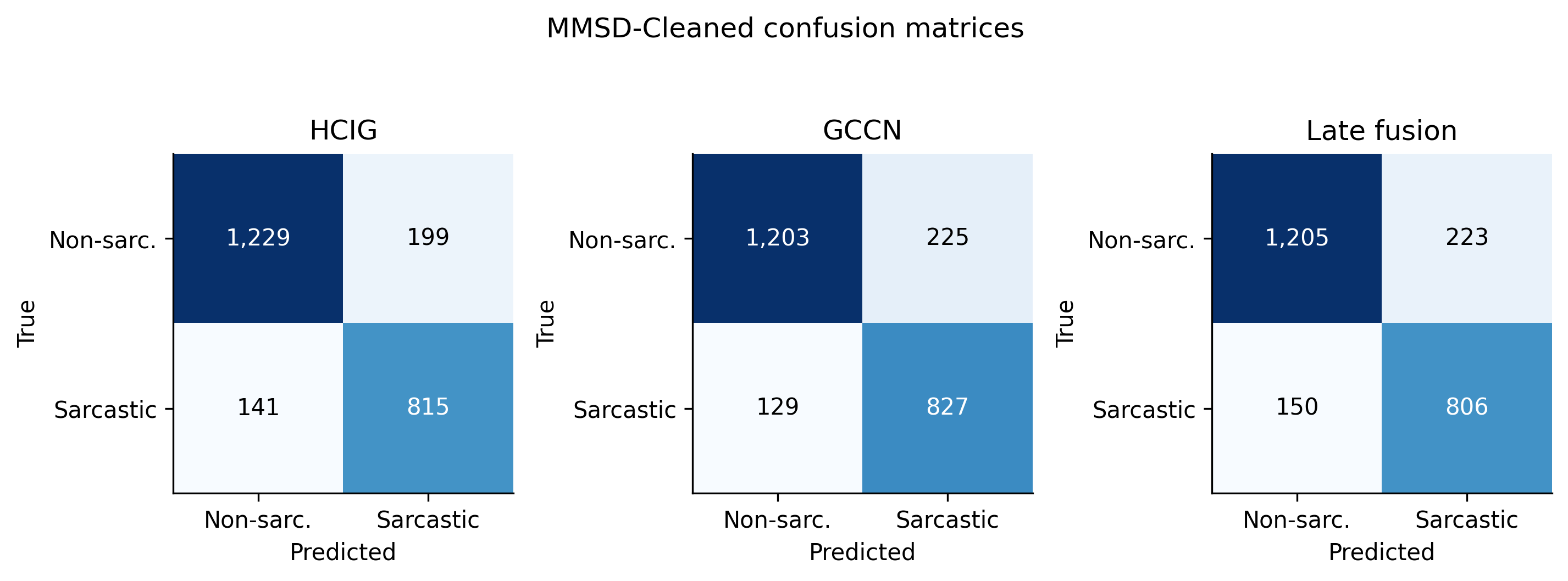}
\caption{Confusion matrices of the evaluated MMSD models.}
\label{fig:confmat}
\end{figure}

\begin{figure}[H]
\centering
\includegraphics[width=\textwidth]{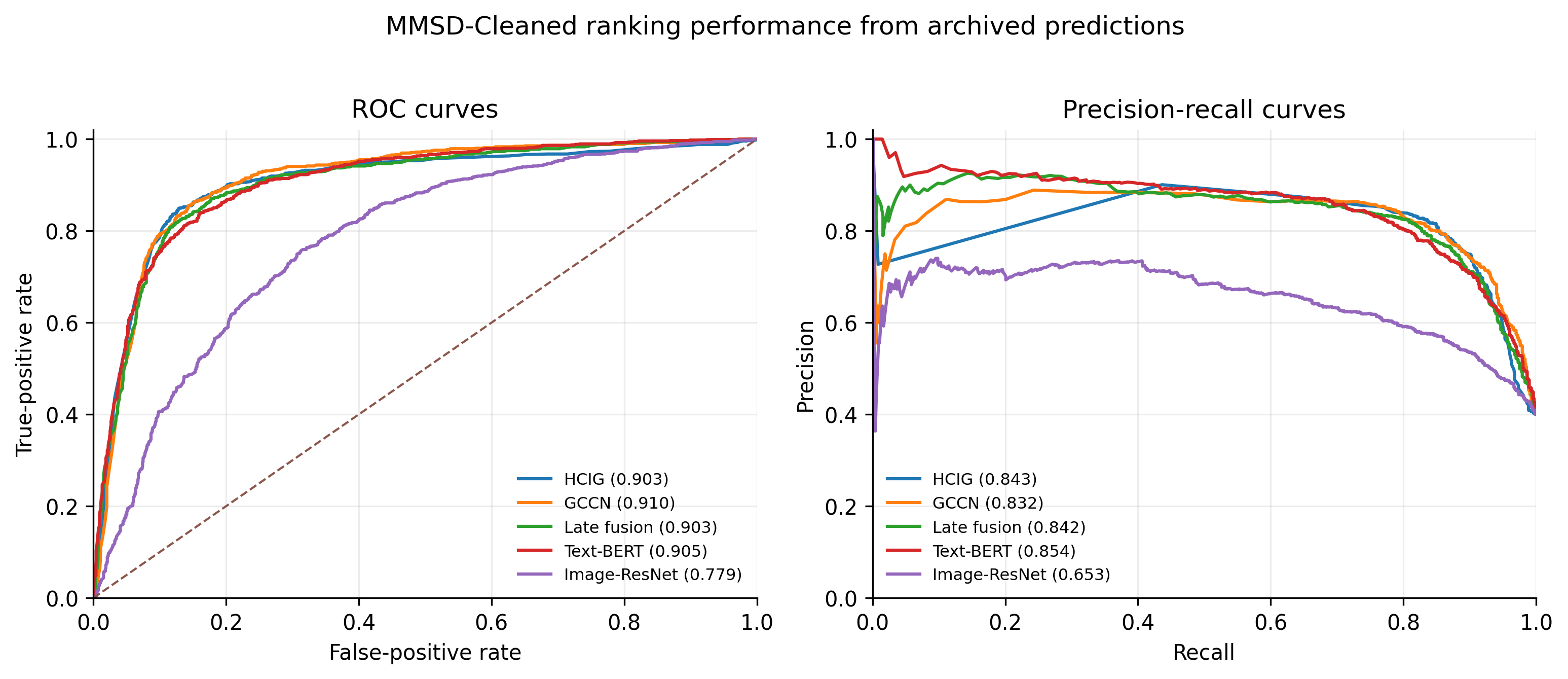}
\caption{ROC curves (left) and precision--recall curves (right) for the MMSD models.}
\label{fig:roc_pr}
\end{figure}

Table~\ref{tab:main} reports test-set performance for all five models on MMSD . HCIG achieves the best accuracy, F1, and macro-F1 of any model; GCCN achieves the best ROC-AUC. Both proposed architectures exceed all three baselines on accuracy, F1, and macro-F1. Figure~\ref{fig:main_bar} visualizes the same comparison across metrics, and Figure~\ref{fig:confmat} shows the corresponding confusion matrices; Figure~\ref{fig:roc_pr} shows ROC and precision--recall curves.

\begin{table}[H]
\centering
\caption{Performance comparison on the MMSD test set.}
\label{tab:main}
\begin{tabular}{lrrrrrrr}
\toprule
Model & Acc. & Prec. & Rec. & F1 & Macro-F1 & AUC & PR-AUC \\
\midrule
Image-ResNet (baseline) & 69.92 & 59.20 & 80.44 & 68.20 & 69.84 & 77.92 & 65.28 \\
Text-BERT (baseline)    & 83.72 & 77.79 & 83.16 & 80.38 & 83.24 & 90.54 & 85.38 \\
Late Fusion (baseline)  & 84.35 & 78.33 & 84.31 & 81.21 & 83.90 & 90.27 & 84.18 \\
GCCN (ours)             & 85.15 & 78.61 & \textbf{86.51} & 82.37 & 84.77 & \textbf{90.99} & 83.19 \\
\textbf{HCIG} (ours)       & \textbf{85.74} & \textbf{80.37} & 85.25 & \textbf{82.74} & \textbf{85.29} & 90.30 & \textbf{84.34} \\
\bottomrule
\end{tabular}
\end{table}

\subsection{Ablation studies}
\label{sec:ablation}
\begin{figure}[H]
\centering
\includegraphics[width=\textwidth]{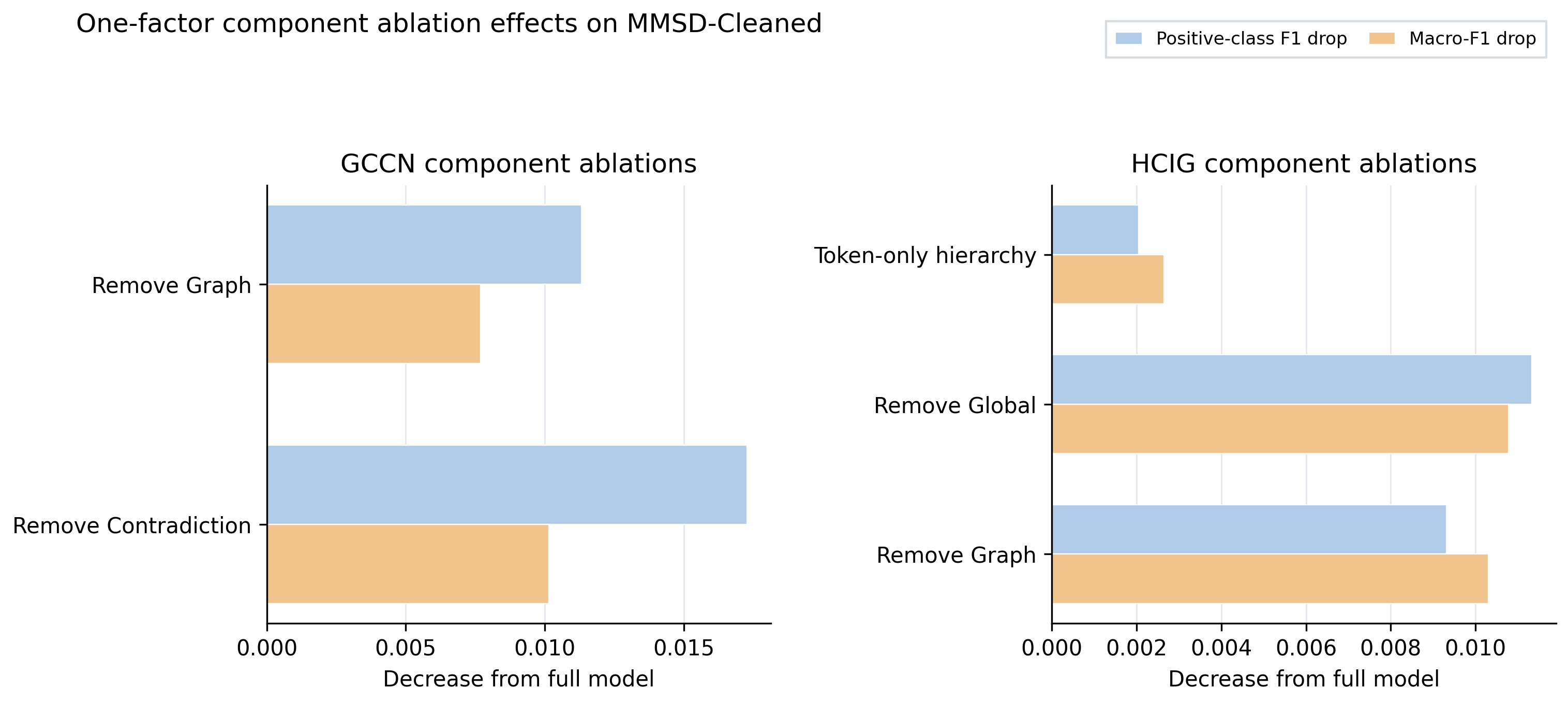}
\caption{Decrease in positive-class F1 and macro-F1 after removing or isolating individual GCCN and HCIG components, relative to each corresponding full model.}
\label{fig:ablation}
\end{figure}

Component ablations were run for GCCN and HCIG on MMSD, holding the training configuration fixed and removing or isolating one architectural component at a time (Tables~\ref{tab:abl_gccn}--\ref{tab:abl_hcig}, Figure~\ref{fig:ablation}). ROC-AUC for the full models is reported only in Table~\ref{tab:main}, which is derived from the canonical main-evaluation prediction archive; the ablation tables are therefore restricted to accuracy and F1-family metrics from the common ablation evaluation.

\begin{table}[H]
\centering
\caption{Ablation study of GCCN on MMSD.}
\label{tab:abl_gccn}
\begin{tabular}{lrrrrr}
\toprule
Variant & Acc. & Prec. & Rec. & F1 & Macro-F1 \\
\midrule
\textbf{GCCN (full)}                    & \textbf{85.15} & \textbf{84.46} & \textbf{85.37} & \textbf{82.37} & \textbf{84.77} \\
w/o graph reasoning             & 84.48 & 83.75 & 84.37 & 81.24 & 84.00 \\
w/o contradiction pooling       & 84.35 & 83.68 & 83.85 & 80.64 & 83.76 \\
text-only (no image branch)     & 84.36 & 83.15 & 84.20 & 81.87 & 84.09 \\
\bottomrule
\end{tabular}
\end{table}

\begin{table}[H]
\centering
\caption{Ablation study of HCIG on MMSD.}
\label{tab:abl_hcig}
\begin{tabular}{lrrrrr}
\toprule
Variant & Acc. & Prec. & Rec. & F1 & Macro-F1 \\
\midrule
\textbf{HCIG (full)}              & \textbf{85.74} & \textbf{85.04} & \textbf{85.66} & \textbf{82.74} & \textbf{85.29} \\
token-level only          & 85.44 & 84.74 & 85.50 & 82.54 & 85.03 \\
w/o global level          & 84.65 & 83.93 & 84.70 & 81.61 & 84.22 \\
w/o graph reasoning       & 84.65 & 83.96 & 84.89 & 81.81 & 84.26 \\
text-only (no image branch)  & 84.23 & 84.15 & 84.83 & 81.88 & 84.74 \\
\bottomrule
\end{tabular}
\end{table}

\subsection{MultiBully in-domain and fine-tuning results}
\begin{figure}[H]
\centering
\includegraphics[width=0.85\textwidth]{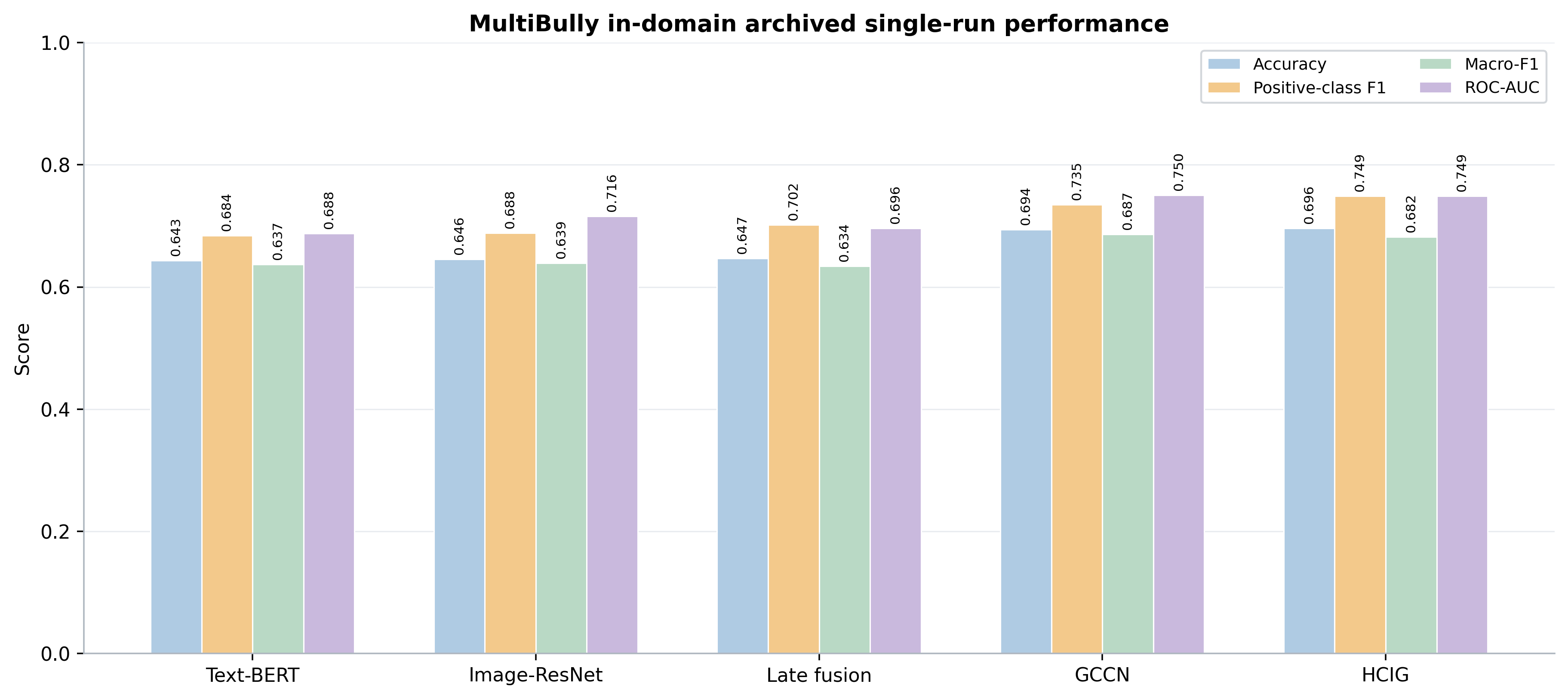}
\caption{MultiBully in-domain comparison across accuracy, positive-class F1, macro-F1, and ROC-AUC.}
\label{fig:mb_eval}
\end{figure}

Table~\ref{tab:multibully} reports MultiBully results under two training regimes: models trained in-domain from randomly initialized heads, and GCCN/HCIG initialized from their complete MMSD checkpoints and then further optimized on MultiBully. The fine-tuning runs retain the existing two-class MMSD classification head and jointly update it with the remaining model parameters; the head is not reinitialized or replaced. Using macro-F1 as the primary class-balanced metric, GCCN obtains the highest in-domain score (68.66\%), whereas HCIG obtains the highest accuracy (69.62\%) and bullying-class F1 (74.90\%). Figure~\ref{fig:mb_eval} shows the in-domain comparison across accuracy, positive-class F1, macro-F1, and ROC-AUC.

\begin{table}[H]
\centering
\caption{Performance comparison on the MultiBully dataset. }
\label{tab:multibully}
\begin{tabular}{llrrrrrr}
\toprule
Model & Regime & Acc. & Prec. & Rec. & F1 & Macro-F1 & AUC \\
\midrule
Text-BERT    & In-domain & 64.33 & 63.88 & 63.66 & 68.43 & 63.71 & 68.78 \\
Image-ResNet & In-domain & 64.56 & 64.11 & 63.83 & 68.83 & 63.88 & 71.60 \\
Late Fusion  & In-domain & 64.67 & 64.36 & 63.46 & 70.17 & 63.43 & 69.64 \\
GCCN         & In-domain & 69.39 & 69.17 & 68.54 & 73.45 & \textbf{68.66} & \textbf{75.02} \\
HCIG         & In-domain & \textbf{69.62} & \textbf{70.08} & 68.20 & \textbf{74.90} & 68.21 & 74.91 \\
\midrule
GCCN   & Fine-tuned from MMSD & 68.24 & 68.54 & 66.80 & 73.76 & 66.77 & 73.18 \\
HCIG   & Fine-tuned from MMSD & 65.59 & 67.35 & 63.25 & 73.47 & 62.27 & 70.73 \\
\bottomrule
\end{tabular}
\end{table}

\subsection{Bidirectional cross-task direct-transfer stress test}
\begin{figure}[H]
\centering
\includegraphics[width=\textwidth]{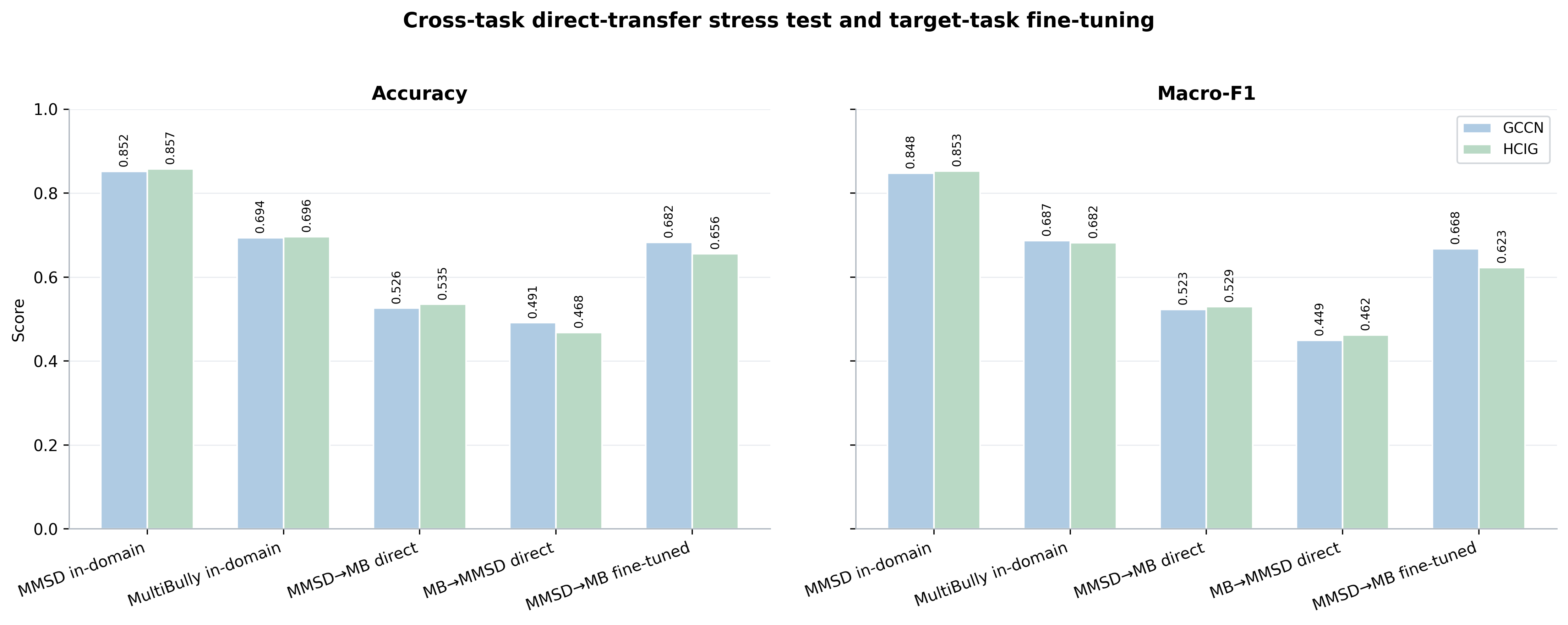}
\caption{Accuracy and macro-F1 for GCCN and HCIG across MMSD in-domain evaluation, MultiBully in-domain evaluation, bidirectional cross-task direct-transfer stress tests, and MMSD-initialized MultiBully fine-tuning.}
\label{fig:zeroshot}
\end{figure}

Models trained exclusively on MMSD were applied directly to the MultiBully test set      (MMSD $\rightarrow$ MB), and models trained exclusively on MultiBully were applied directly to the MMSD test set (MB$\rightarrow$MMSD), in both cases without target-task training (Figure~\ref{fig:zeroshot}). Because MMSD predicts sarcasm whereas MultiBully predicts bullying, the source and target labels represent different constructs; this experiment is therefore treated as a bidirectional cross-task direct-transfer stress test rather than conventional zero-shot domain transfer. For both GCCN and HCIG, direct-transfer accuracy in both directions falls in the 51--56\% range, close to the corresponding majority-class baseline, and neither architecture shows a clear advantage under this cross-task classifier reuse. The in-domain differences between GCCN and HCIG (Tables~\ref{tab:main} and~\ref{tab:multibully}) are not preserved in the stress-test setting.

\subsection{Explainability and qualitative analysis}
\begin{figure}[H]
\centering
\includegraphics[width=0.75\textwidth]{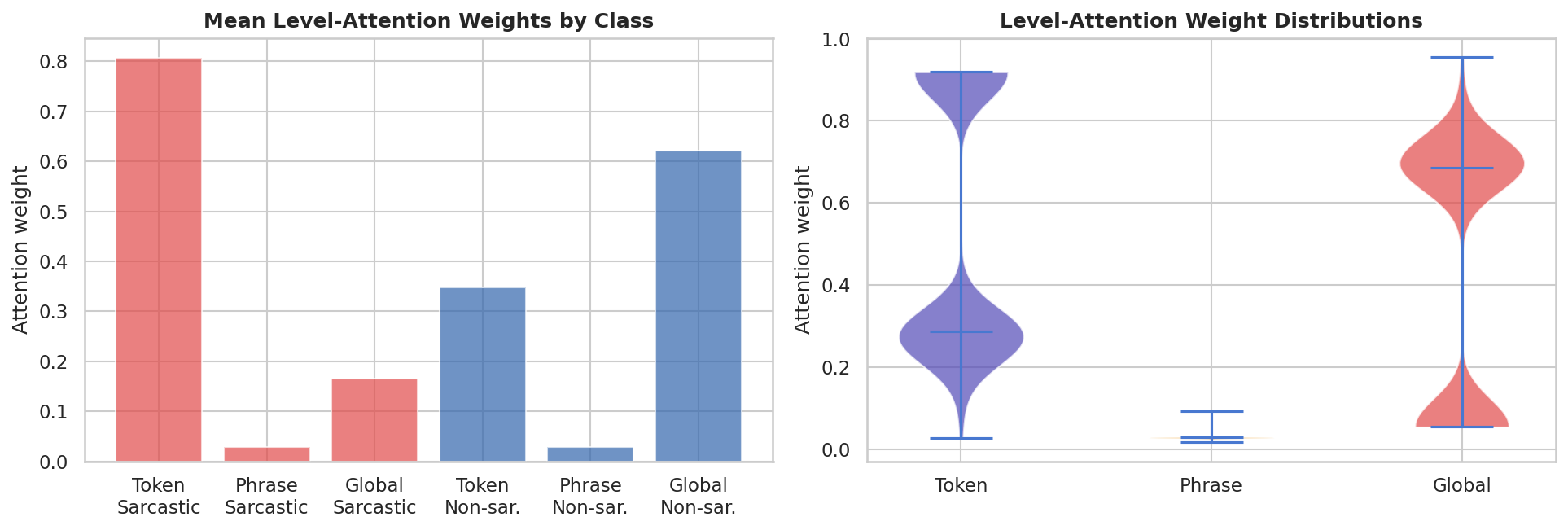}
\caption{HCIG level-attention weights (token/phrase/global) by label.}
\label{fig:hcig_attn}
\end{figure}

\begin{figure}[H]
\centering
\includegraphics[width=0.75\textwidth]{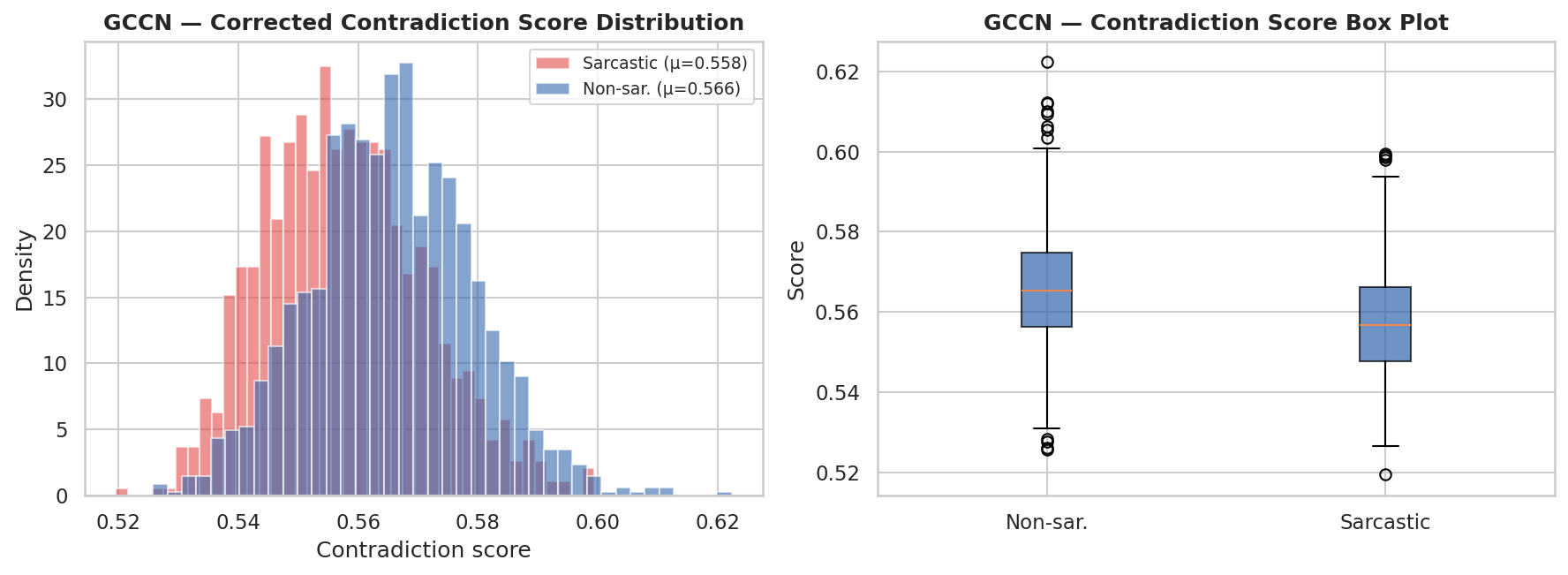}
\caption{GCCN contradiction score by label.}
\label{fig:gccn_contra}
\end{figure}

Figure~\ref{fig:hcig_attn} shows HCIG's learned level-attention weights (token/phrase/global) by label, and Figure~\ref{fig:gccn_contra} shows GCCN's contradiction score distribution by label. HCIG's level-attention weights show a strong and consistent preference for the token-level branch across both classes, with the global branch contributing a secondary signal and the phrase-level branch contributing comparatively little, consistent with the ablation finding (Table~\ref{tab:abl_hcig}) that a token-only variant already recovers most of HCIG's full performance.

\section{Conclusion and Future Work}
\label{sec:conclusion}
This paper presented HCIG, a novel Hierarchical Cross-modal Incongruity Graph Network for multimodal sarcasm and cyberbullying detection by explicitly modeling semantic incongruity between textual and visual modalities at token, phrase, and global levels. Unlike conventional multimodal fusion approaches that primarily focus on feature aggregation, HCIG performs hierarchical graph-based reasoning to capture fine-grained cross-modal inconsistencies and adaptively integrates multi-level representations through an attention-guided fusion mechanism. In addition, a lightweight Graph-based Cross-modal Contradiction Network (GCCN) was introduced as a complementary architecture to investigate contradiction-aware multimodal reasoning. Extensive experiments on the MMSD and MultiBully benchmark datasets demonstrate the effectiveness of the proposed frameworks. HCIG achieved the best performance on MMSD with an accuracy of 85.74\% and a macro-F1 score of 85.29\%, whereas GCCN obtained the highest macro-F1 score (68.66\%) on MultiBully while HCIG achieved the highest accuracy (69.62\%) and bullying-class F1 (74.90\%). Furthermore, the ablation studies demonstrate the effectiveness of hierarchical incongruity modeling, highlighting that token-level semantic interactions provide the strongest discriminative cues while phrase-level and global representations contribute complementary contextual information. The cross-task transfer analysis further indicates that sarcasm and cyberbullying exhibit distinct semantic characteristics, emphasizing the importance of task-specific multimodal reasoning for effective social media understanding.

The rapid advancement of multimodal artificial intelligence and foundation models is expected to drive the next generation of multimodal content understanding systems. Future research should focus on developing more context-aware frameworks capable of reasoning over implicit sarcasm, offensive intent, cultural references, conversational history, and temporal dependencies across multiple modalities. Integrating graph reasoning with large language models and large multimodal models offers promising opportunities for improving semantic alignment, cross-modal reasoning, and generalization across diverse social media domains. Future benchmark datasets should also include multilingual, multicultural, and real-world multimodal content while adopting standardized annotation protocols and evaluation settings to improve reproducibility and facilitate meaningful comparison across studies. In addition, enhancing explainability, adversarial robustness, fairness, computational efficiency, and human-centered AI will be essential for building reliable and trustworthy multimodal moderation systems. We believe that the proposed HCIG framework provides a strong foundation for future research toward robust, interpretable, and scalable multimodal sarcasm and cyberbullying detection systems.

\section*{Data and Code Availability}
All result tables reported in this paper are provided as supplementary CSV files. MMSD is available from \cite{cai2019multimodal}; MultiBully is available from \cite{maity2022multitask}.

\bibliographystyle{ieeetr}
\bibliography{references}

\end{document}